\documentclass{article}
\usepackage{arxiv}
\renewcommand{\shorttitle}{Learning to Plan by Looking Back}

\usepackage[utf8]{inputenc}
\usepackage[T1]{fontenc}
\usepackage{microtype}
\usepackage{amsmath,amssymb,amsthm}
\usepackage[numbers,square,sort&compress]{natbib}

\usepackage[ruled,vlined]{algorithm2e}
\DontPrintSemicolon
\SetKwInOut{Input}{Input}
\SetKwInOut{Output}{Output}
\SetKwInOut{Given}{Given}

\usepackage{tikz}
\usepackage{graphicx}
\usetikzlibrary{arrows.meta}

\usepackage[hidelinks]{hyperref}

\usepackage{xcolor}

\theoremstyle{plain}
\newtheorem*{problem}{The Problem}
\newtheorem*{remark}{Remark}

\newcommand{\Verify}{\mathsf{Verify}}
\newcommand{\X}{\mathcal{X}}
\newcommand{\Y}{\mathcal{Y}}
\newcommand{\Z}{\mathcal{Z}}
\newcommand{\Loss}{\mathcal{L}}
\newcommand{\Transition}{\mathcal{T}}
\newcommand{\Actions}{\mathcal{A}}
\newcommand{\States}{\bar{\mathcal{S}}}
\newcommand{\Batch}{\mathsf{SampleBatch}}
\newcommand{\Penalty}{\mathsf{Penalty}}
\newcommand{\Utility}{\mathsf{Utility}}
\newcommand{\Train}{\mathsf{Train}}
\newcommand{\DisSol}{\mathsf{DiscoveredSolution}}
\newcommand{\Score}{\mathsf{Score}}
\newcommand{\ConstructTriple}{\mathsf{ConstructTriple}}
\newcommand{\length}{\mathsf{len}}
\newcommand{\fail}{\mathsf{fail}}

\DeclareMathOperator*{\argmax}{arg\,max}
\DeclareMathOperator*{\argmin}{arg\,min}

\hypersetup{
  pdfauthor={Lars Simon, Holger Eble, Manuel Radons}
}

\title{Learning to Plan by Looking Back:\\
Hindsight Hierarchies for Training Reasoning Models}

\author{
  Lars Simon\\
  Bundesdruckerei GmbH\\
  Berlin, Germany\\
  \texttt{lars.simon@bdr.de}
  \And
  Holger Eble\\
  Bundesdruckerei GmbH\\
  Berlin, Germany\\
  \texttt{holger.eble@bdr.de}
  \And
  Manuel Radons\\
  Bundesdruckerei GmbH\\
  Berlin, Germany\\
  \texttt{manuel.radons@bdr.de}
}
\date{}

\begin{document}
\maketitle

\begin{abstract}
\phantomsection
\label{sec:abstract}
\noindent
We introduce a self-improvement loop for reasoning models based on the following observation:
Even when the difficulty of a  problem exceeds the model's current solving abilities, an additionally supplied solution might enable the model to extract useful solution ideas in hindsight.
We operationalize this by jointly training the same model to exhibit the following three capabilities: predicting solution ideas from problems alone, reverse-engineering ideas from problems and known solutions, and solving problems using provided ideas.
The loop alternates between reverse engineering such ideas from problems with supplied solutions and using these ideas as additional supervision for joint training of all three capabilities.
We give a formal specification of our method and a concrete instantiation for interactive theorem proving in the Lean theorem prover; empirical evaluation remains future work.
\end{abstract}

\section{Introduction}
\label{sec:introduction}
Our method is inspired by the following asymmetry in problem solving: It is, in general, harder to solve a problem from scratch than to reverse engineer the idea underlying a known solution.
For a reasoning model, this presents a learning opportunity: even when a problem exceeds its current solving abilities, a supplied solution may enable it to extract a useful idea for solving that problem in hindsight.
Such a reconstructed idea can then provide additional supervision for learning to develop a solution idea from the problem alone, before attempting to solve it.
The opportunity is therefore to \emph{learn from ideas that the model can already reverse engineer, but may not yet be able to discover unaided}.

We operationalize this observation by jointly training a model
on three tasks.
A \emph{hindsight component} reverse engineers solution ideas
from problems and known solutions.
A \emph{foresight component} predicts such ideas from problems
alone.
A \emph{solver component} uses problems and provided ideas to construct solution candidates.
All three components share a backbone network; for example, when their inputs and outputs can be represented as text, we may use a shared language model with role-specific
prompt templates.

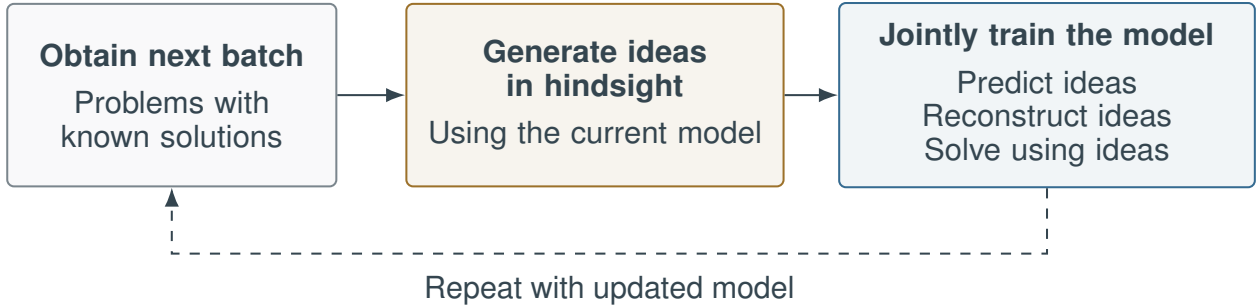
\begin{figure}[tbp]
\centering
\begingroup
\definecolor{overviewBlue}{HTML}{356B91}
\definecolor{overviewOchre}{HTML}{9C712A}
\definecolor{overviewInk}{HTML}{34454F}
\resizebox{\linewidth}{!}{%
\begin{tikzpicture}[
  >=Latex,
  box/.style={
    draw=overviewInk!65,fill=overviewInk!3,
    rounded corners=2pt,align=center,
    font=\sffamily\footnotesize,text=overviewInk,
    inner sep=6pt,minimum height=1.9cm,line width=.6pt
  },
  flow/.style={->,semithick,draw=overviewInk}
]

\node[box,text width=3.0cm] (data) at (0,0)
  {\textbf{Obtain next batch}\\[4pt]
   Problems with\\known solutions};

\node[box,draw=overviewOchre,fill=overviewOchre!8,
      text width=3.5cm] (hindsight) at (4.4,0)
  {\textbf{Generate ideas\\
  in hindsight}\\[4pt]
   Using the current model};

\node[box,draw=overviewBlue,fill=overviewBlue!7,
      text width=3.9cm] (train) at (9.1,0)
  {\textbf{Jointly train the model}\\[4pt]
   Predict ideas\\
   Reconstruct ideas\\
   Solve using ideas};

\draw[flow] (data.east) -- (hindsight.west);
\draw[flow] (hindsight.east) -- (train.west);

\draw[flow,dashed] (train.south) -- (9.1,-1.65)
  -- node[midway,below=3pt,font=\sffamily\footnotesize,
          text=overviewInk]
     {Repeat with updated model}
  (0,-1.65) -- (data.south);

\end{tikzpicture}%
}
\endgroup
\caption{Simplified schematic of the training loop. Each iteration generates hindsight ideas for a new batch and jointly trains the same model to predict, reconstruct, and use ideas.}
\label{fig:training-overview}
\end{figure}

The centerpiece of our method is an alternating training loop.
In each iteration, the current hindsight component reconstructs ideas for a batch of problems with supplied solutions.
We then jointly train all three components using the resulting data together with previously collected examples.
Concretely, given a problem $x$ with solution $y$ and reverse-engineered idea $z$, the foresight component is trained to predict $z$ given $x$, the hindsight component is trained to predict $z$ given $x$ and $y$, and the solver component is trained to predict $y$ given $x$ and $z$.
The next iteration uses the updated hindsight component to generate ideas.
Thus, the capability responsible for producing additional supervision is itself an explicit training target.
Figure \ref{fig:training-overview} illustrates the central idea.

We express solution ideas at several levels of detail, forming a \emph{hierarchy} from overarching strategies to detailed plans.
Hindsight starts from a known solution, reconstructs a detailed plan, and compresses it into coarser ideas.
Foresight proceeds in the opposite direction, predicting a coarse idea from the problem and progressively refining it.
During inference, the solver uses the hierarchy predicted by the foresight component to construct a solution candidate.
Figure \ref{fig:hierarchy_directions} illustrates these directions and the use of hierarchies during inference.
The ideas reverse-engineered by the hindsight component are intended to provide guidance
that could plausibly be followed from the problem alone, rather than merely describe the supplied solution or reflect \emph{hindsight bias}.
In our Lean instantiation, we seek to avoid the latter through explicit instructions in the language model's prompt templates.

One possible weakness of our method lies in the fact that the hindsight component is trained on its own outputs.
While this might be partially mitigated by the fact that the three components share a backbone network, it would be desirable to select among reverse-engineered hierarchies in order to introduce an additional training signal for the hindsight component in the spirit of expert iteration \citep{expertiteration2017}.
We provide an optional refinement of our method which accomplishes this in the case of \emph{verifiable problem domains}, i.e., when proposed solution candidates can be procedurally verified:
For each problem and reference solution, we generate several
candidate hierarchies in hindsight.
The solver then repeatedly attempts the problem using each hierarchy, while the reference solution is withheld.
We then compute a score for each hierarchy, based on its success ratio under the repeated solver attempts and, optionally, domain-specific quantities.
We retain the highest-scoring hierarchy and an associated valid solution for training, provided the score meets an acceptance threshold.

We provide a formal specification of our method for a general problem setting, together with the optional refinement for verifiable problem domains.
We also give a concrete instantiation for interactive theorem proving in the Lean theorem prover \citep{lean42021}, where natural language proof ideas guide tactic generation during proof search.
The method is implemented in an accompanying software project; empirical evaluation remains future work.

The paper is organized as follows.
Section \ref{sec:related_work} discusses related work.
Section \ref{sec:problem} introduces the problem setting, and Section \ref{sec:method} specifies the general method.
Section \ref{sec:concrete_lean_instantiation} presents the Lean instantiation, with implementation details in Appendix \ref{sec:appendix}.
Section \ref{sec:discussion} discusses limitations and directions for empirical evaluation.

\section{Related work}
\label{sec:related_work}

\paragraph{Language models for interactive theorem proving.}
Language models have been used for proof-step generation and guided proof search \citep{gptf2020,pact2022,htps2022}.
Further work incorporates premise retrieval \citep{leandojo2023} and integration with automated theorem provers \citep{thor2022}.
Recent Lean provers also explore large-scale synthetic proof data and reinforcement learning from proof assistant feedback \citep{deepseekprover2024,deepseekprover15_2025,
deepseekprover2_2025,goedelprover2025}.
Our Lean instantiation builds on language-model-guided proof search, conditioning tactic generation on learned hierarchies of proof ideas.

\paragraph{Planning and hierarchical reasoning.}
Explicit planning and problem decomposition organize reasoning with language models into intermediate steps and subproblems \citep{planandsolve2023,leasttomost2023,decomposedprompting2023},
while tree-based search explores alternative reasoning paths
\citep{treeofthoughts2023}.
In formal theorem proving, informal proofs can guide the construction of formal proof skeletons whose gaps are filled by an automated prover \citep{dsp2023}.
Modular and recursive approaches construct proofs through intermediate lemmas and subgoals \citep{legoprover2024,poetry2024,deepseekprover2_2025}.
Our hierarchies are intended to describe solution ideas at different levels of granularity, providing guidance for subsequent solving.

\paragraph{Iterative self-training and verifier-guided selection.}
Expert iteration uses a learned policy to guide search and then trains that policy to imitate the decisions produced by search \citep{expertiteration2017}, leading to an iterative improvement loop.
Related self-training methods learn from selected model-generated solutions \citep{selfimprove2023,restem2024}.
In formal theorem proving, systems interleave proof search with training on discovered proofs \citep{formalcurriculum2023,htps2022}, while self-play approaches additionally generate new conjectures \citep{stp2025}.
Learned correctness estimators, trained on final outcomes or individual reasoning steps, also support candidate selection at inference time \citep{verifiers2021,verifystep2024}.
In the case of verifiable problem domains, our method uses separate solving attempts to assess reverse-engineered hierarchies and selects training triples based on that assessment.

\paragraph{Retrospective supervision and hierarchical guidance.}
STaR iteratively trains on self-generated rationales, including rationales constructed retrospectively using correct-answer hints that are removed before fine-tuning \citep{star2022}.
On-policy self-distillation transfers next-token distributions conditioned on a reference solution to predictions made without access to that solution \citep{opsd2026}.
Lean-STaR uses a teacher language model to retrospectively annotate proof states and known next tactics with natural language thoughts.
It then trains a model to generate thoughts and tactics, followed by expert iteration on its own successful proof trajectories \citep{leanstar2025}.
DeepInsight uses an external language model to extract core techniques and proof sketches from reference proofs, and trains a model through progressive multi-stage supervised fine-tuning and hierarchy-aware policy optimization \citep{deepinsight2026}.

Our method makes retrospective generation of hierarchies (i.e., rationales at several levels of detail) an explicitly trained task.
A shared backbone is \emph{jointly} trained on three tasks:
reverse engineering hierarchies from problems and reference solutions; predicting such hierarchies from problems alone; and solving problems using these hierarchies.
Each retained training triple supervises all three tasks.
The training loop alternates between generating training triples by reverse engineering hierarchies and jointly training on the above three tasks.
Consequently, subsequent retrospective hierarchy generation uses the updated model, with the aim of improving the hierarchies available for further training.
This differs from the forward generation of thoughts and tactics in Lean-STaR's expert iterations, and from DeepInsight's retrospective rationale generation using a teacher model:
our loop repeatedly performs retrospective hierarchy generation with the same backbone that learns to predict and use the resulting hierarchies.
Finally, in verifiable problem domains, we additionally assess the reverse-engineered hierarchies through separate solving attempts, in which the model receives the problem and hierarchy while the reference solution is withheld.
We then select training triples based on this assessment, introducing an additional training signal through data selection, in the spirit of expert iteration \citep{expertiteration2017}.

\section{Problem setting}
\label{sec:problem}

Let $\X$ be a set of \emph{problem instances} and let $\Y$ be a set of \emph{candidate solutions}. Write $\mathcal R\subseteq\X\times\Y$ for the relation specifying valid solutions. A problem instance can admit no solution, exactly one solution, or more than one solution. We do not initially assume that membership in $\mathcal R$ can be decided by an available algorithm.
\begin{problem}
Given $x\in\X$, construct $y\in\Y$ such that $(x,y)\in\mathcal R$.
\end{problem}
Training starts from a collection of supplied pairs $(x,y)\in\X\times\Y$,
where $y$ is a reference solution for $x$. The idealized description assumes
that these are valid solutions, i.e., $(x,y)\in\mathcal R$. The training procedure can still
be executed with imperfect reference solutions, but -- without a verifier -- the quality of these 
reference solutions is an assumption about the source of the data.

\subsection{Verifiable problem domains}
\label{sec:verifiable_domains}

In a \emph{verifiable problem domain}, we assume access to a deterministic verification procedure
\begin{align*}
    \Verify \colon \X\times\Y \to \{0,1\},
\end{align*}
which, on every input $(x,y)$ we consider, terminates and returns $1$ if and only if $(x,y)\in\mathcal R$, i.e., if and only if $y$ is a valid solution for $x$. This allows us to repeatedly query $\Verify$ in our algorithm, see Section \ref{sec:method}.

\begin{remark}
By a slight abuse of notation, $\Verify$ refers to both a procedure and the set-theoretic map computed by this procedure, which is just the indicator function $\mathbf{1}_\mathcal{R}$ of $\mathcal R$ on $\X\times\Y$. Since $\mathbf{1}_\mathcal{R}$ is guaranteed to exist without further assumptions on the problem domain, the key requirement is access to a deterministic procedure for evaluating $\mathbf{1}_\mathcal{R}$ that terminates on all considered inputs.
\end{remark}

\paragraph{Example: interactive theorem proving.}
We focus on interactive theorem provers that support step-by-step execution of proof commands (tactics), such as Lean. From a current proof state with at least one open goal a single tactic is provided to the interactive theorem prover, which either responds with a successor state or signals failure. 

Let $\mathcal{S}$ denote the set of valid (i.e., accepted by the prover) proof states with at least one open goal and let $\Actions$ denote the set of tactics\footnote{We restrict $\Actions$ to tactics that do not admit unproved obligations or introduce disallowed axioms. In Lean, this excludes, for example, \texttt{sorry} and \texttt{admit}, including their use within other tactics.}.
For the sake of simplicity, we identify all valid proof states with no open goals and denote them as $\top$. Similarly, we identify all error states and denote them as $\bot$. Consequently, the set of all states is $\States = \mathcal{S} \cup \{ \top , \bot \}$.

The interactive theorem prover then provides a transition function\footnote{As before, we assume access to a deterministic procedure for evaluating this function that terminates on all inputs we consider. In Lean, tactic execution need not terminate; even kernel type checking can fail to terminate on certain contrived inputs. As a practical fallback, we can impose timeouts in the implementation, treating a timeout as rejection.}
\begin{align*}
    \Transition\colon \States \times \Actions \to \States ,
    \qquad
    \Transition (\top,a)=\Transition (\bot,a)=\bot \text{ for all }a\in\Actions ,
\end{align*}
where $\Transition (s,a)$ denotes the successor state when tactic $a\in\Actions$ is executed in state $s\in\States$.
For $s\in\mathcal{S} , a\in\Actions$, execution of tactic $a$ in state $s$
either results in an error (iff $\Transition (s,a) = \bot$) or produces a valid successor state; in the latter case, $\Transition (s,a) = \top$ iff no goals remain, otherwise $\Transition (s,a) \in \mathcal{S}$.

We now instantiate interactive theorem proving as a special case of verifiable problem domains: Letting
\begin{align*}
    \X := \mathcal{S} ,
    \qquad
    \Y := \bigcup_{k=1}^\infty \Actions^k ,
\end{align*}
we define $\Verify\colon\X\times\Y\to\{0,1\}$ by imposing that, for $s_0\in \mathcal{S}, (a_1 , \dots , a_k )\in\Actions^k$, we have $\Verify (s_0, (a_1 , \dots , a_k )) = 1$ iff $s_k = \top$, where $s_{j+1}=\Transition (s_j , a_{j+1})$ for $j=0,\dots , k-1$.

\section{Method}
\label{sec:method}

Building on the key ideas introduced in Section \ref{sec:introduction}, we now give a formal description of our method.
We first describe the shared architecture and training loop, parameterized by a procedure for constructing hindsight training triples.
We then specify two such procedures: one applicable to the general problem setting introduced in Section \ref{sec:problem}, and one tailored to the verifiable problem domains introduced in Section \ref{sec:verifiable_domains}.
We deliberately leave domain-specific choices open in this section; Section \ref{sec:concrete_lean_instantiation} provides a concrete instantiation of our method for interactive theorem proving in Lean.

\subsection{Hierarchies}
\label{sec:hierarchies}

With $\X$ and $\Y$ as in Section \ref{sec:problem}, we fix a depth $n\in\mathbb{Z}_{\geq 0}$, and introduce \emph{hierarchy sets} $\Z_0 , \dots , \Z_n$. An element $z=(z_0 , \dots , z_n )\in \Z :=\Z_0 \times \dots \times \Z_n$ will be called a \emph{hierarchy}. In practice, a hierarchy $z\in\Z$ for a problem $x\in \X$ is intended to represent a \emph{semantically meaningful} idea, plan, or approach for constructing a solution $y\in \Y$ for $x$ at different levels of granularity: From a detailed solution sketch $z_0$ to a coarse description of the overarching idea $z_n$. How the hierarchy sets are chosen and how the different levels of granularity are encouraged are problem-dependent; our choices in the special case of interactive theorem proving in Lean are detailed in Sections \ref{sec:lean_hierarchies_and_data} and \ref{sec:lean_ConstructTriple}, respectively.

\subsection{System components}
\label{sec:components}

The system, parametrized by a collection of parameters $\theta$, consists of three components. While these components may have component-specific parameters, they share a backbone network which accounts for the majority of parameters (note that $\theta$ denotes the collection of all parameters: shared backbone and, possibly, component-specific parameters). The three components are
\begin{itemize}
    \item a \textbf{solver component} $S_\theta$ which, given a problem $x\in\X$ and a hierarchy $z\in\Z$, produces a solution candidate $\hat{y}\in\Y$; we write $\hat{y}\sim S_\theta (\cdot \mid x, z)$,
    \item a \textbf{foresight component} $F_\theta$ which, given a problem $x\in\X$, iteratively predicts a hierarchy $\hat{z}\in\Z$ from coarsest to finest; we write $\hat{z}\sim F_\theta (\cdot \mid x)$. More precisely, $F_\theta$ is defined by a family $\left(F_{j,\theta}\right)_{j=0,\dots ,n}$ via $\hat{z}_n\sim F_{n,\theta}(\cdot\mid x)$ and $\hat{z}_j\sim F_{j,\theta}(\cdot\mid x,\hat{z}_{j+1})$ for all $j\in\{0, \dots , n-1\}$,
    \item a \textbf{hindsight component} $H_\theta$ which, given a problem $x\in\X$ with solution $y\in\Y$, iteratively reverse engineers a hierarchy $\tilde{z}\in\Z$ from finest to coarsest; we write $\tilde{z}\sim H_\theta (\cdot \mid x,y)$. More precisely, $H_\theta$ is defined by a family $\left(H_{j,\theta}\right)_{j=0,\dots ,n}$ via $\tilde{z}_0\sim H_{0,\theta}(\cdot\mid x,y)$ and $\tilde{z}_j\sim H_{j,\theta}(\cdot\mid x,\tilde{z}_{j-1})$ for all $j\in\{1, \dots , n\}$.
\end{itemize}
Figure \ref{fig:hierarchy_directions} illustrates the opposite directions of hindsight and foresight, and how the solver uses the predicted hierarchy during inference, for a simple mathematical example.
\begin{figure}[tbp]
\centering
\resizebox{\linewidth}{!}{%
\begin{tikzpicture}[
  x=1cm,y=1cm,>=Latex,
  box/.style={
    draw=black!55,rounded corners=2pt,align=center,
    font=\sffamily\footnotesize,inner sep=5pt,
    minimum height=1.55cm
  },
  input/.style={box,text width=1.85cm,fill=black!3},
  plan/.style={box,text width=3.10cm},
  output/.style={box,text width=2.05cm},
  flow/.style={->,semithick,draw=black!75},
  tag/.style={
    font=\sffamily\scriptsize,align=center,
    fill=none,inner sep=1pt
  }
]

\node[font=\sffamily\footnotesize,align=center]
  at (7.05,1.55)
  {\textbf{Illustrative problem $x$:}
   prove $\sum_{i=1}^{m} i=m(m+1)/2$
   for $m\in\mathbb Z_{\geq1}$.};

% (a) Hindsight.
\node[font=\sffamily\footnotesize\bfseries,anchor=west]
  at (0,0.85)
  {(a) Hindsight: reverse engineer detailed idea, then compress};

\node[input] (hin) at (1.10,-0.25)
  {\textbf{Given}\\Problem $x$\\Solution $y$};

\node[plan,fill=orange!12] (hd) at (4.80,-0.25)
  {\textbf{Detailed sketch $\tilde z_0$}\\
   Check $m=1$; assume case $m$;
   add $m+1$ and simplify.};

\node[plan,fill=orange!12] (hc) at (8.95,-0.25)
  {\textbf{Coarse idea $\tilde z_1$}\\Induction on $m$.};

\node[output,fill=black!3] (hout) at (13.00,-0.25)
  {\textbf{Hierarchy}\\$\tilde z=(\tilde z_0,\tilde z_1)$};

\draw[flow] (hin) --
  node[tag,above=4pt] {$H_{0,\theta}$} (hd);
\draw[flow] (hd) --
  node[tag,above=4pt] {$H_{1,\theta}$} (hc);
\draw[flow] (hc) -- (hout);

% (b) Foresight.
\node[font=\sffamily\footnotesize\bfseries,anchor=west]
  at (0,-1.40)
  {(b) Foresight: predict coarse idea, then refine};

\node[input] (fin) at (1.10,-2.60)
  {\textbf{Given}\\Problem $x$ only};

\node[plan,fill=blue!8] (fc) at (4.80,-2.60)
  {\textbf{Coarse idea $\hat z_1$}\\Induction on $m$.};

\node[plan,fill=blue!8] (fd) at (8.95,-2.60)
  {\textbf{Detailed sketch $\hat z_0$}\\
   Check $m=1$; assume case $m$;
   add $m+1$ and simplify.};

\node[output,fill=black!3] (fout) at (13.00,-2.60)
  {\textbf{Hierarchy}\\$\hat z=(\hat z_0,\hat z_1)$};

\draw[flow] (fin) --
  node[tag,above=4pt] {$F_{1,\theta}$} (fc);
\draw[flow] (fc) --
  node[tag,above=4pt] {$F_{0,\theta}$} (fd);
\draw[flow] (fd) -- (fout);

% (c) Inference.
\node[font=\sffamily\footnotesize\bfseries,anchor=west]
  at (0,-3.75)
  {(c) Inference: generate a hierarchy in foresight, then solve};

\node[input] (iin) at (1.10,-4.95)
  {\textbf{Given}\\Problem $x$ only};

\node[plan,fill=blue!8] (ihier) at (4.80,-4.95)
  {\textbf{Predicted hierarchy}\\$\hat z=(\hat z_0,\hat z_1)$};

\node[plan,fill=green!9] (isolve) at (8.95,-4.95)
  {\textbf{Solver $S_\theta$}\\Input: $x,\hat z$};

\node[output,fill=green!9] (iout) at (13.00,-4.95)
  {\textbf{Solution candidate}\\$y'$};

\draw[flow] (iin) --
  node[tag,above=4pt] {$F_\theta$} (ihier);
\draw[flow] (ihier) -- (isolve);
\draw[flow] (isolve) -- (iout);

\node[font=\sffamily\scriptsize,align=center]
  at (7.05,-6.10)
  {All foresight and hindsight predictions are conditioned on $x$. Hindsight, foresight, and solving share model parameters $\theta$.};

\end{tikzpicture}%
}
\caption{Hierarchy generation in both directions (two levels, $n=1$)
and inference. The solver receives the full predicted hierarchy.
The example texts are hand-written illustrations,
not model outputs or empirical observations.}
\label{fig:hierarchy_directions}
\end{figure}

In practice, the system may be implemented using a multi-headed neural network, or (in the case where elements of $\X$, $\Y$, and $\Z$ can be represented by text) using a language model with role-specific prompt templates or special tokens indicating the three roles, e.g., $\langle\textsc{solve}\rangle$, $\langle\textsc{fore}\rangle$, and $\langle\textsc{hind}\rangle$. Our choices in the special case of interactive theorem proving in Lean are detailed in Section \ref{sec:lean_system_components}.

Note that conditioning the three components on $x$ is necessary: For example, in the context of interactive theorem proving, assume the idea at the coarsest hierarchy level $z_n$ is ``induction on vector space dimension''. Of course, for $n\geq 1$, it is not possible for the foresight component to refine this to a detailed solution sketch $z_0$ without knowledge of the problem $x$ to be solved.

In interactive theorem proving, the solver component $S_\theta$ denotes the entire search procedure, including both tactic generation and proof search.
This procedure may return a failure outcome instead of a finite tactic sequence.
Formally, we can accommodate this by adjoining the failure outcome to $\Y$ and declaring it invalid for every problem instance.

\paragraph{Inference.} We describe the inference procedure of our system for fixed parameters $\theta$:
Given a problem instance $x\in\X$, we first generate a hierarchy $\hat{z}\sim F_\theta (\cdot \mid x)$ using the foresight component.
Subsequently, we use the solver component to produce a solution candidate $y'\sim S_\theta (\cdot \mid x, \hat{z})$.

In the \emph{general problem setting}, when no verifier is available, we simply return $y'$.

In a \emph{verifiable problem domain}, we evaluate $\Verify (x,y')$ to check whether $y'$ solves $x$; in case of success we return $y'$, else we signal failure.  At test time, we may increase success probability (at the cost of additional compute) by running a nested loop: we sample multiple hierarchies in an outer loop and, for each hierarchy, sample multiple solution candidates in an inner loop; verification is performed immediately after each candidate is generated, and the procedure terminates as soon as a solution for $x$ is found (or a predetermined budget is exhausted).

\subsection{Ingredients for training the system}
\label{sec:ingredients}

We require four more ingredients before describing the full training loop, namely a \emph{seed dataset}, a way to  \emph{obtain additional data}, a way to \emph{construct training triples in hindsight}, and a \emph{training operator} for training the three components on datasets consisting of triples in $\X\times\Y\times\Z$. Here we give an abstract description of these four ingredients and refer to Section \ref{sec:concrete_lean_instantiation} for our concrete choices in the special case of interactive theorem proving in Lean. The procedures introduced in this subsection are not assumed to be deterministic; they may be randomized or approximate.

\paragraph{Seed dataset.} We assume access to a \emph{seed dataset} $D_0$ of triples in $\X\times\Y\times\Z$. For each entry $(x,y,z)$ of $D_0$, $y$ is a reference solution for $x$ and $z$ is intended to be a hierarchy for constructing $y$ from $x$ in the abstract sense explained in Section \ref{sec:hierarchies}; in verifiable problem domains, we moreover require that $\Verify (x,y)=1$. In practice, most available datasets consist of pairs in $\X\times\Y$, making it necessary to augment data points of the form $(x,y)$ with \emph{seed hierarchies}. How the latter can be obtained is problem-dependent; in the case where elements of $\X$, $\Y$, and $\Z$ can be represented by text, an existing language model can be used to reverse engineer seed hierarchies for such data points. The seed dataset may also be empty, e.g., when a pretrained model already provides a suitable initialization for the three components.

\paragraph{Obtaining additional data.}
The full training loop requires additional data beyond the seed
dataset $D_0$, sourced from an external dataset or generated
synthetically. Crucially, we only need data points of the form
$(x,y)\in\X\times\Y$, where $y$ is a reference solution for $x$,
since the hindsight component will be used to reverse engineer
hierarchies for such data points. In verifiable problem domains,
we additionally require that $\Verify(x,y)=1$ for every such data point.

We assume access to a procedure which, in iteration $t$, returns a batch $\Batch(t)$ of such pairs. This procedure may depend on, e.g., the current model parameters $\theta$, previously collected data, external datasets, synthetic data generators, and resource
budgets; such dependencies are suppressed in the notation.

In verifiable problem domains, one optional refinement is to enforce that $\Batch (t)$ has prescribed proportions of problems from different difficulty groups.
These difficulty groups can be defined by observed success under different inference budgets, using the system with the current parameters $\theta$ and a specified evaluation protocol.
This includes ``difficult, but doable'' problems that are only solved under a large budget but not under an ordinary budget; problems not solved under any tested budget may form a separate difficulty group.

\paragraph{Constructing training triples in hindsight.} We assume access to a procedure $\ConstructTriple$ which, given a pair $(x,y)$ with reference solution $y$ for $x$ (satisfying $\Verify(x,y)=1$ in verifiable problem domains), returns either a triple $(x,\bar{y}, z)$ or a distinguished symbol $\fail$, indicating that no suitable training triple was obtained:
\begin{align*}
    \ConstructTriple (x,y;\theta ) \in 
    (\{x\}\times\Y\times\Z)
    \cup\{\fail\}
\end{align*}
Here $z$ is a hierarchy generated by the current hindsight component $H_\theta (\cdot | x,y)$ and $\bar{y}$ is a reference solution for $x$ which, in general, can be different from $y$. In verifiable problem domains, we additionally require that $\Verify(x,\bar{y})=1$.

\emph{The procedure $\ConstructTriple$ encapsulates the main difference between the two versions of our method for the general problem setting and for verifiable problem domains:}
In the former, we simply query the current hindsight component, obtaining $z\sim H_\theta (\cdot | x,y)$, and then return $(x,y,z)$.
However, in verifiable problem domains, access to the procedure $\Verify$ allows us to assess the ``usefulness'' of various generated hierarchies for the solver component $S_\theta$; consequently, by selecting hierarchies based on their estimated usefulness, we introduce an additional training signal, similar in spirit to expert iteration \citep{expertiteration2017}.
This signal enters entirely through data selection; the training objective remains unchanged.
We describe these two choices for $\ConstructTriple$ in detail in Sections \ref{sec:construct_hierarchies_general} and \ref{sec:construct_hierarchies_verifiable}, respectively.

\paragraph{Training operator.}  We assume access to a training procedure $\Train$, which, given the current parameters $\theta$ and a dataset $D$ of triples in $\X\times\Y\times\Z$, returns updated parameters $\theta_{\operatorname{new}} = \Train (D; \theta)$. The procedure will be used on datasets $D$ with entries of the form $(x,y,z)$, where $y$ is a reference solution for $x$ and, ideally, realizes the hierarchy $z$. Consequently, the training objective of the procedure is chosen so that $S_\theta (\cdot \mid x, z)$, $F_\theta (\cdot \mid x)$, and $H_\theta (\cdot \mid x, y)$ are encouraged to predict $y$, $z$, and $z$ respectively. As the three components share a backbone network, each of the individual training signals affects all three components. On empty datasets, $\Train$ leaves the parameters unchanged. Our concrete choice of $\Train$ in the case of interactive theorem proving in Lean is described in Section \ref{sec:lean_train}.

\subsection{The full training loop}
\label{sec:full-training-loop}

We start by training the system once on the seed dataset, setting $\theta \leftarrow \Train(D_0;\theta_{\mathrm{init}})$, where $\theta_{\mathrm{init}}$ denotes the initial parameters.
Subsequently, we execute a loop that alternates between obtaining a batch $\Batch(t)$ of pairs in $\X\times\Y$, constructing training triples in $\X\times\Y\times\Z$ from these pairs in hindsight, and training the system on a dataset composed of these newly constructed triples and previously collected data. Figure \ref{fig:training-loop} illustrates the training loop including the two choices for $\ConstructTriple$ from Sections \ref{sec:construct_hierarchies_general} and \ref{sec:construct_hierarchies_verifiable}; Algorithm \ref{alg:training-loop} gives the pseudocode.

\begin{figure}[tbp]
\centering
\begingroup
\definecolor{loopBlue}{HTML}{356B91}
\definecolor{loopGreen}{HTML}{35745E}
\definecolor{loopOchre}{HTML}{9C712A}
\definecolor{loopInk}{HTML}{34454F}
\resizebox{\linewidth}{!}{%
\begin{tikzpicture}[
  x=1cm,y=1cm,>=Latex,
  box/.style={
    draw=loopInk!65,fill=loopInk!3,rounded corners=2pt,
    align=center,font=\sffamily\footnotesize,
    inner sep=5pt,line width=.6pt
  },
  flow/.style={->,semithick,draw=loopInk},
  heading/.style={
    font=\sffamily\footnotesize\bfseries,
    text=loopInk,anchor=west
  },
  note/.style={
    font=\sffamily\scriptsize,text=loopInk,align=center
  }
]

% Initialization and full training loop.
\node[heading] at (0,1.40)
  {(a) Full training loop};

\node[box,draw=loopBlue,fill=loopBlue!7,
      text width=5.5cm,minimum height=.85cm]
  (seed) at (3.0,.45)
  {\textbf{Seed initialization}\\
   Train on seed dataset of triples $(x,y,z)$.};

\node[heading] at (6.4,-.20)
  {Repeat for $t=1,\ldots,T$};

\node[box,text width=3.05cm,minimum height=2.05cm]
  (batch) at (1.9,-1.65)
  {\textbf{Obtain batch of}\\
   $(x,y)$ \textbf{pairs}\\[3pt]
   $B\leftarrow\Batch(t)$};

\node[box,draw=loopOchre,fill=loopOchre!8,
      text width=3.65cm,minimum height=2.05cm]
  (construct) at (6.0,-1.65)
  {\textbf{For each $(x,y)$ in $B$}\\[3pt]
   \textbf{construct triple $(x,\bar{y},z)$}\\[3pt]
   \textbf{in hindsight, or $\fail$}\\[3pt]
   $\ConstructTriple(x,y;\theta)$};

\node[box,text width=3.35cm,minimum height=2.05cm]
  (replay) at (10.15,-1.65)
  {\textbf{Build dataset}\\[3pt]
   Replay policy:\\[3pt]
   new triples {\scriptsize (discard $\fail$)}\\
   + earlier triples};

\node[box,draw=loopBlue,fill=loopBlue!7,
      text width=3.25cm,minimum height=2.05cm]
  (train) at (14.25,-1.65)
  {\textbf{Joint training}\\[3pt]
   $S_\theta,\quad F_\theta,\quad H_\theta$};

\draw[flow] (seed.south) -- (3.0,-.35) -- (1.9,-.35)
  -- (batch.north);
\draw[flow] (batch) -- (construct);
\draw[flow] (construct) -- (replay);
\draw[flow] (replay) -- (train);

\draw[flow,dashed] (train.south) -- (14.25,-2.95)
  -- (1.9,-2.95) -- (batch.south);

\node[note,fill=white,inner sep=3pt] at (8.1,-2.95)
  {Next iteration with updated parameters $\theta$};

\node[heading] at (0,-3.65)
  {(b) Two choices for $\ConstructTriple(x,y;\theta)$};

% Direct augmentation.
\draw[draw=loopBlue!55,fill=loopBlue!3,rounded corners=3pt]
  (0,-4.0) rectangle (16.2,-5.85);

\node[heading,text=loopBlue] at (.25,-4.38)
  {Direct augmentation with hindsight hierarchies};

\node[note,anchor=east] at (15.95,-4.38)
  {No verifier required};

\node[box,draw=loopBlue!65,fill=white,
      text width=5.6cm,minimum height=.85cm]
  (direct) at (4.05,-5.15)
  {\textbf{Generate a hierarchy in hindsight}\\
   $z\sim H_\theta(\cdot\mid x,y)$};

\node[box,draw=loopBlue!65,fill=white,
      text width=5.6cm,minimum height=.85cm]
  (directreturn) at (12.15,-5.15)
  {\textbf{Return the training triple}\\
   $(x,y,z)$};

\draw[flow,draw=loopBlue] (direct) -- (directreturn);

% Verifier-guided selection.
\draw[draw=loopGreen!55,fill=loopGreen!3,rounded corners=3pt]
  (0,-6.20) rectangle (16.2,-10.15);

\node[heading,text=loopGreen] at (.25,-6.58)
  {Verifier-guided selection of hindsight hierarchies};

\node[note,anchor=east] at (15.95,-6.58)
  {$\Verify(x,y)=1$};

\node[box,draw=loopGreen!65,fill=white,
      text width=3.4cm,minimum height=2.85cm]
  (candidates) at (2.1,-8.45)
  {\textbf{Generate $K$}\\
   \textbf{hindsight hierarchies}\\[5pt]
   $\tilde z^{(i)}\sim H_\theta(\cdot\mid x,y)$\\[3pt]
   $i=1,\ldots,K$};

\node[box,draw=loopGreen!65,fill=white,
      text width=5.1cm,minimum height=2.85cm]
  (assess) at (7.75,-8.45)
  {\textbf{Assess each hierarchy}\\
   {\scriptsize Example implementation}\\[2pt]
   Sample several solution candidates\\
   $\sim S_\theta (\cdot | x,\tilde z^{(i)})$; check using $\Verify$.\\[2pt]
   Compute $\Score(i)$ from outcomes\\
   and domain-specific quantities.\\[2pt]
   Retain valid $\tilde y^{(i)}$ (fallback: $y$).};

\node[box,draw=loopGreen!65,fill=white,
      text width=4.45cm,minimum height=2.85cm]
  (select) at (13.7,-8.45)
  {\textbf{Select hierarchy $i^\star$}\\[4pt]
   Highest score:\\
   $i^\star\in\argmax_i \Score (i)$\\[5pt]
   \textbf{Return selected triple}\\
   $(x,\tilde y^{(i^\star)},\tilde z^{(i^\star)})$\\[3pt]
   if $\Score(i^\star)\geq\delta$; else $\fail$.};

\draw[flow,draw=loopGreen] (candidates) -- (assess);
\draw[flow,draw=loopGreen] (assess) -- (select);

\end{tikzpicture}%
}
\endgroup
\caption{Schematic for the full training loop and the two choices
for $\ConstructTriple$ from Sections
\ref{sec:construct_hierarchies_general} and
\ref{sec:construct_hierarchies_verifiable}.
The assessment box illustrates one possible way to score candidate hierarchies and retain a valid solution.}
\label{fig:training-loop}
\end{figure}

Specifically, in iteration $t=1,\dots , T$, we obtain a batch $B\leftarrow\Batch(t)$ and initialize an empty dataset $D_t$. 
For each entry $(x,y)$ of $B$, we invoke $\ConstructTriple(x,y;\theta)$ and append the returned triple to $D_t$, unless the procedure returns $\fail$. The parameters $\theta$ remain fixed throughout this construction phase. 
After processing the batch, we form a training dataset $D$ from the new dataset $D_t$ and previously collected data from $D_0,\dots,D_{t-1}$.
This is accomplished using a \emph{replay policy} which specifies how new and previously collected entries are mixed.
Subsequently, we update the parameters via $\theta\leftarrow\Train(D;\theta)$.

In practice, the procedures $\ConstructTriple$ and $\Train$ may depend on the iteration $t$.
This dependence on $t$ is suppressed in the notation used throughout this section.

\begin{algorithm}[t]
\small
\Given{Iteration count $T\in\mathbb{Z}_{\geq1}$;
procedures $\Batch$, $\ConstructTriple$, and $\Train$;
replay policy}
\Input{Seed dataset $D_0$;
initial parameters $\theta_{\mathrm{init}}$}
\Output{Trained parameters $\theta$}

$\theta\leftarrow\Train(D_0;\theta_{\mathrm{init}})$\;

\For{$t=1,\ldots,T$}{
    $B\leftarrow\Batch(t)$\;
    Initialize an empty dataset $D_t$\;

    \ForEach{entry $(x,y)$ of $B$}{
        $r\leftarrow\ConstructTriple(x,y;\theta)$
        \tcp*{triple with a hindsight hierarchy, or $\fail$}
        \If{$r\neq\fail$}{
            Append the triple $r$ to $D_t$\;
        }
    }

    Form a training dataset $D$ from $D_t$ and $D_0,\dots,D_{t-1}$ according to the replay policy\;

    $\theta\leftarrow\Train(D;\theta)$\;
}

\Return{$\theta$}\;

\caption{The full training loop.}
\label{alg:training-loop}
\end{algorithm}

\subsection{$\ConstructTriple$ in the general problem setting: direct augmentation with hindsight hierarchies}
\label{sec:construct_hierarchies_general}

In the general problem setting, see Section \ref{sec:problem}, we construct a training triple directly from a supplied pair $(x,y)\in\X\times\Y$, where $y$ is a reference solution for $x$. Concretely, we sample a hierarchy $z$ in hindsight using the current parameters $\theta$, and then augment the pair $(x,y)$ with $z$:
\[
\begin{array}{@{}l@{}}
\ConstructTriple(x,y;\theta):\\[2pt]
\qquad \textbf{sample } z\sim H_\theta(\cdot\mid x,y)\\
\qquad \textbf{return } (x,y,z)
\end{array}
\]
We refer to this as \emph{direct augmentation with hindsight hierarchies}. Note that this can also be used in a verifiable problem domain when verifier-guided selection (see Section \ref{sec:construct_hierarchies_verifiable}) is not desired.

\subsection{$\ConstructTriple$ in verifiable problem domains: verifier-guided selection of hindsight hierarchies}
\label{sec:construct_hierarchies_verifiable}

We now describe our choice of $\ConstructTriple$ in verifiable problem domains, see Section \ref{sec:verifiable_domains}. In this setting, the supplied pairs $(x,y)\in\X\times\Y$ are assumed to satisfy $\Verify (x,y)=1$.
A priori, a hierarchy $z$ sampled from $H_\theta (\cdot \mid x,y)$ need not be \emph{useful}, in the sense that the solver component, given $(x,z)$, is likely to discover a solution for $x$. We therefore sample several hindsight-derived hierarchies and select among them using an estimate of their usefulness, together with a penalty encouraging the intended levels of granularity. We refer to this as \emph{verifier-guided selection of hindsight hierarchies}.
Algorithm \ref{alg:verifier_guided_hindsight_hierarchy_selection} summarizes the procedure.
We describe its ingredients and selection rule below.

\paragraph{Assessing usefulness and retaining a reference solution.} We assume access to a procedure which, given $(x,y,z)\in\X\times\Y\times\Z$ with $\Verify (x,y)=1$, jointly returns a nonnegative usefulness estimate $\Utility (x,y,z;\theta)$ and a valid solution $\DisSol (x,y,z;\theta)$ for $x$, i.e.,
\[
    \Verify\bigl(x,\DisSol(x,y,z;\theta)\bigr)=1.
\]
Larger values of $\Utility$ are intended to indicate higher estimated usefulness of the hierarchy $z$ for the solver, given problem $x$. The solution returned by $\DisSol$ need not equal $y$; this is intentional: if the hierarchy $z$ is useful for the solver, then $S_\theta (\cdot |x,z)$ is likely to discover \emph{some} valid solution for $x$, which need not be equal to or even ``similar to'' $y$. Accordingly, $\DisSol (x,y,z;\theta)$ is intended to be a valid solution for $x$ that the solver is likely to discover under the hierarchy $z$. We write $(\Utility, \DisSol )(x,y,z;\theta)$ for the output of this joint procedure.

A natural implementation of this joint procedure samples several solution candidates from $S_\theta(\cdot\mid x,z)$ and checks them using $\Verify$.
The outcome of this process provides a usefulness estimate, based on the success ratio and optionally domain-specific quantities, while a valid solution found in this process can be retained as $\DisSol(x,y,z;\theta)$.
If none is found, we retain $y$ instead. Our concrete choice for the joint procedure in the case of interactive theorem proving in Lean is described in Section \ref{sec:lean_ConstructTriple}.

\paragraph{Encouraging different levels of granularity.}
To encourage the generated hierarchies to represent ideas at different levels of granularity (as opposed to, for example, listing several ideas at the same level of detail), we assume access to a procedure, which, given $z\in\Z$, returns a non-negative real number $\Penalty (z)$, which penalizes departure from this requirement.
Such a penalty can encourage compression (between hierarchy levels), but does not by itself establish that a hierarchy is semantically meaningful.
Our concrete choice of $\Penalty$ in the case of interactive theorem proving in Lean is described in Section \ref{sec:lean_ConstructTriple}.

\paragraph{Selecting a training triple.}
Given $(x,y)\in \X\times\Y$ with $\Verify (x,y)=1$, we sample $K\in\mathbb{Z}_{\geq 1}$ candidate hierarchies  $\tilde{z}^{(i)}\sim H_\theta (\cdot \mid x,y)$, where $i=1,\dots ,K$. Each candidate hierarchy is assessed by jointly obtaining a usefulness estimate and a valid solution for $x$ (recall that $\DisSol$ always has the valid solution $y$ available as fallback),
\begin{align*}
                    ( u^{(i)} , \tilde{y}^{(i)} )
                    \leftarrow
                    (\Utility , \DisSol ) (x,y,\tilde{z}^{(i)};\theta)
                    .
\end{align*}
We combine the usefulness estimate with the penalty to obtain a score; subsequently, we choose a hierarchy maximizing the score (ties are broken by candidate order):
\begin{align*}
 \Score(i)=u^{(i)}-\lambda\cdot\Penalty(\tilde z^{(i)}),
 \qquad \lambda>0,
 \qquad i^\star\in\argmax_{i\in\{1,\ldots,K\}}\Score(i).
\end{align*}
If $\Score (i^\star )\geq \delta$ for a prescribed threshold $\delta >0$, we let $\ConstructTriple$ return the triple $(x, \tilde{y}^{(i^\star)} , \tilde{z}^{(i^\star)} )$, else we return $\fail$.
Algorithm \ref{alg:verifier_guided_hindsight_hierarchy_selection} summarizes this specification of $\ConstructTriple$.

\begin{algorithm}[t]
\small
\Given{Hindsight component $H_\theta$, candidate count $K\in\mathbb{Z}_{\geq1}$;
joint procedure $(\Utility,\DisSol)$;
penalty $\Penalty$; hyperparameters $\lambda,\delta>0$}
\Input{Pair $(x,y)$ with $\Verify(x,y)=1$;
parameters $\theta$, held fixed throughout this call}
\Output{One training triple in $\{x\}\times\Y\times\Z$ or $\fail$}
\For{$i=1,\ldots,K$}{
 Sample $\tilde z^{(i)}\sim H_\theta(\cdot\mid x,y)$\;
 $(u^{(i)},\tilde y^{(i)})\leftarrow
 (\Utility,\DisSol)(x,y,\tilde z^{(i)};\theta)$\;
 $\Score(i)\leftarrow u^{(i)}-\lambda\cdot\Penalty(\tilde z^{(i)})$\;
}
$i^\star\leftarrow\argmax_{i\in\{1,\ldots,K\}}\Score(i)$
\tcp*{ties broken by candidate order}
\If{$\Score(i^\star)\geq\delta$}{
 \Return{$(x,\tilde y^{(i^\star)},\tilde z^{(i^\star)})$}\;
}
\Return{$\fail$}\;
\caption{$\ConstructTriple$ in verifiable problem domains: verifier-guided selection of hindsight hierarchies.}
\label{alg:verifier_guided_hindsight_hierarchy_selection}
\end{algorithm}

\section{A concrete Lean instantiation}
\label{sec:concrete_lean_instantiation}

To instantiate our method in the special case of interactive theorem proving (see Section \ref{sec:verifiable_domains}), we provide concrete re\-al\-i\-za\-tions of the abstract concepts introduced in Section \ref{sec:method}. We use Lean and the verifier-guided choice for $\ConstructTriple$ from Section \ref{sec:construct_hierarchies_verifiable}, as summarized in Algorithm \ref{alg:verifier_guided_hindsight_hierarchy_selection}.
We use a single language model as the shared backbone of all three system components, with role-specific prompt templates.
We specify the Lean instantiation of our method below; Appendix \ref{sec:appendix} provides concrete implementation details, including hyperparameter settings and prompt templates.

\subsection{Lean instantiation: hierarchies and data}
\label{sec:lean_hierarchies_and_data}

We first specify the hierarchy representation and then describe how to construct the seed dataset $D_0$ of triples and the procedure $\Batch$ for obtaining additional batches of pairs in the context of interactive theorem proving in Lean.

\paragraph{Hierarchies.}
Formally, $\Z_0 = \dots = \Z_n$ is realized as the set of strings over an appropriate alphabet, and we impose no further constraints.
However, we aim to ensure during training that the hierarchies represent semantically meaningful proof ideas at different levels of granularity.
This motivates the construction of the seed dataset described below and the choice of $\Penalty$, see Section \ref{sec:lean_ConstructTriple}. The concrete prompt templates in Appendix \ref{sec:appendix_prompt_templates} use two levels ($n=1$): a detailed plan $z_0$ and a coarse proof idea $z_1$; neither is required to consist of executable Lean commands.

\paragraph{Seed dataset.} Starting from a dataset of pairs $(x,y)= (s_0 , (a_1 , \dots , a_k )) \in \X\times\Y$ with $\Verify (x,y)=1$, we use a \emph{teacher language model} to reverse engineer natural language seed hierarchies for these pairs; the prompt template for this teacher includes instructions to avoid \emph{hindsight bias}. We then augment the pairs $(x,y)$ with the computed hierarchies, yielding the seed dataset $D_0$.

Importantly, the teacher is only used for the generation of the seed dataset, i.e., does not appear in the subsequent training loop, where hindsight hierarchies are exclusively generated by the hindsight component. Our concrete choice for the initial dataset of pairs and for the teacher language model is described in Appendix \ref{sec:appendix_data}.

Alternatively, if our chosen language model has been pretrained on technical material and already produces useful hindsight hierarchies for some problems, $D_0$ may be empty, avoiding the need for a teacher to generate seed hierarchies.

\paragraph{Obtaining additional data.} We assume access to a supply of pairs $(x,y)\in \X\times\Y$ with $\Verify (x,y)=1$; in practice, these are either sourced from an external dataset or generated synthetically. Ordinary sampling from such a supply, or generating pairs on demand, suffices to obtain a procedure $\Batch$ satisfying the specifications from Section \ref{sec:ingredients}. As an optional refinement, one may prescribe proportions of problems from different difficulty groups (relative to the current parameters $\theta$ of the system) as described in Section \ref{sec:ingredients}. For example, there may be groups consisting of problems that are solved under an ordinary proof search budget, problems that are solved under a generous (but not under an ordinary) proof search budget, and problems that are not solved under any of the tested proof search budgets. Our concrete choice of $\Batch$ is given in Appendix \ref{sec:appendix_data}.

\subsection{Lean instantiation: system components}
\label{sec:lean_system_components}
% describe:
%      foresight component
%      hindsight component
%      solver component (including proof search algorithm)

All three components are implemented using a single underlying language model with shared parameters, and we distinguish between components using role-specific prompt templates.
We implement the foresight component $F_\theta$ and the hindsight component $H_\theta$ via iterative calls to the language model that mirror the setup in Section \ref{sec:components}.
The solver component $S_\theta$ executes a best-first search, using the language model as a tactic oracle.
We give the details below and refer to Appendix \ref{sec:appendix_prompt_templates} for the concrete prompt templates.

\paragraph{Foresight component.}
For $F_\theta$ we first query the language model with a text representation of the proof state $s_0\in\X = \mathcal{S}$ and a prompt requesting a coarse proof idea. We then iteratively refine the idea by querying the language model with the same text representation of $s_0$, with the idea output in the previous iteration, and with a prompt requesting a more detailed idea.

\paragraph{Hindsight component.}
The hindsight component $H_\theta$ is implemented analogously: In the first step, given $(x,y) = (s_0 , (a_1 , \dots , a_k )) \in \X\times \Y$ with $\Verify (x,y) = 1$, we query the language model with a text representation of $s_0$, with a text representation of the tactic sequence $(a_1 , \dots , a_k )$, and with a prompt requesting a reverse-engineered detailed proof idea.
Optionally, if the context window of the language model allows, we may also include the intermediate proof states $s_1 , \dots , s_{k-1}$ in the prompt, which can be obtained by querying the transition function $\Transition$ of the interactive theorem prover.
We then iteratively compress the idea by querying the language model with the same text representation of $s_0$, with the idea output in the previous iteration, and with a prompt requesting a coarser idea. All prompts for the hindsight component contain instructions to avoid hindsight bias and to produce proof ideas that could plausibly be followed from the initial state, rather than merely describe the supplied proof.

\paragraph{Solver component.}
Given a problem $s_0\in \X = \mathcal{S}$ and a hierarchy $z\in\Z$, the solver component $S_\theta$ performs best-first proof search starting from $s_0$, using the shared language model as a tactic oracle, see, e.g., \citep{leandojo2023}.
When expanding a proof state $s$, we query the language model with text representations of the initial proof state $s_0$, the hierarchy $z$, and the current proof state $s$, requesting a next tactic.
The hierarchy $z$ and the initial state $s_0$ remain fixed throughout the search; they are nevertheless included in every prompt to the language model (together with the state at the current node in the proof search graph): the purpose of the hierarchy is to guide tactic generation, and the initial state provides context for the hierarchy. Our configuration of best-first proof search, including the priority rule, tactic generation settings, and search budgets, is specified in Appendix \ref{sec:appendix_best_first_search}.

Unlike an executed tactic, a hierarchy receives no immediate feedback from Lean.
During proof search, we therefore treat the hierarchy as guidance that may be incomplete or partly wrong.
Our solver prompt encourages the language model to use the hierarchy where it fits the current proof state, while allowing it to depart from the suggested plan when necessary.
The concrete prompt template is provided in Appendix \ref{sec:appendix_prompt_templates}.

\paragraph{Inference.}
For fixed parameters $\theta$, we follow the inference procedure described in Section \ref{sec:components}:
Given a problem $s_0\in \X = \mathcal{S}$, we first generate a hierarchy $\hat{z}\sim F_\theta(\cdot \mid s_0)$ using the foresight component, and then invoke the solver component $S_\theta (\cdot \mid s_0 , \hat{z})$.
We refer to this procedure as an \emph{arm}.
In order to explore several hierarchies, we may execute several arms sequentially, each with a newly sampled hierarchy and a fresh proof search graph rooted at $s_0$.
If an arm finds no solution, we proceed to the next arm.
This procedure stops as soon as a valid solution is found, or signals failure when all arms have been exhausted.
Implementation details, such as the concrete number of arms, together with their search budgets, are provided in Appendix \ref{sec:appendix_multiple_arms}, while Figure \ref{fig:multi_arm_lean_search} illustrates this procedure.

\begin{figure}[tbp]
\centering
\begingroup
\definecolor{leanBlue}{HTML}{356B91}
\definecolor{leanGreen}{HTML}{35745E}
\definecolor{leanOchre}{HTML}{9C712A}
\definecolor{leanInk}{HTML}{34454F}
\resizebox{\linewidth}{!}{%
\begin{tikzpicture}[
  x=1cm,y=1cm,>=Latex,
  box/.style={
    draw=leanInk!65,fill=white,rounded corners=2pt,
    align=center,font=\sffamily\footnotesize,
    inner sep=5pt,line width=.6pt
  },
  flow/.style={->,semithick,draw=leanInk},
  heading/.style={
    font=\sffamily\footnotesize\bfseries,
    text=leanInk,anchor=west
  },
  note/.style={
    font=\sffamily\scriptsize,text=leanInk,align=center
  }
]

% Panel backgrounds and headings.
\draw[draw=leanBlue!50,fill=leanBlue!3,rounded corners=3pt]
  (0,0) rectangle (4.10,8.25);

\draw[draw=leanGreen!50,fill=leanGreen!3,rounded corners=3pt]
  (4.50,0) rectangle (16.20,8.25);

\node[heading] at (.25,7.88)
  {(a) Sequential arms};

\node[heading] at (4.75,7.88)
  {(b) Search within one arm};

% Left panel: sequential arms.
\node[box,draw=leanBlue,fill=leanBlue!8,
      text width=3.10cm,minimum height=1.20cm]
  (armone) at (2.05,6.65)
  {\textbf{Arm $1$}\\[3pt]
   $\hat z^{(1)}\sim F_\theta(\cdot\mid s_0)$\\
   Search from $s_0$.};

\node[box,draw=leanBlue,fill=leanBlue!5,
      text width=3.10cm,minimum height=1.20cm]
  (armtwo) at (2.05,4.60)
  {\textbf{Arm $2$}\\[3pt]
   $\hat z^{(2)}\sim F_\theta(\cdot\mid s_0)$\\
   New search from $s_0$.};

\node[inner sep=0pt,minimum width=1mm,minimum height=3mm]
  (armdots) at (2.05,3.15) {};

\foreach \offset in {-1,0,1}{
  \fill[leanInk]
    ([yshift=\offset mm]armdots.center) circle (.45pt);
}

\node[box,draw=leanBlue,fill=leanBlue!5,
      text width=3.10cm,minimum height=1.20cm]
  (armlast) at (2.05,1.70)
  {\textbf{Arm $J$}\\[3pt]
   $\hat z^{(J)}\sim F_\theta(\cdot\mid s_0)$\\
   New search from $s_0$.};

\draw[flow] (armone.south) --
  node[note,right=3pt] {no solution}
  (armtwo.north);

\draw[flow] (armtwo.south) --
  ([yshift=1mm]armdots.north);

\draw[flow] ([yshift=-1mm]armdots.south) --
  (armlast.north);

\node[note,text width=3.45cm] at (2.05,.40)
  {Stop at the first valid solution.};

% Right panel: fixed inputs and initialization.
\node[box,draw=leanOchre,fill=leanOchre!10,
      text width=4.40cm,minimum height=1.15cm]
  (fixed) at (7.50,7.00)
  {\textbf{Fixed throughout this arm}\\
   Initial state $s_0$ and hierarchy $\hat z$};

\node[box,fill=leanInk!3,
      text width=4.65cm,minimum height=1.15cm]
  (initialize) at (13.30,7.00)
  {\textbf{Initialize proof search graph}\\
   Rooted at $s_0$.};

% Shared dotted connector; the left-panel boxes stay in place.
\coordinate (armjoin) at (4.30,7.00);

\draw[semithick,draw=leanInk,densely dotted]
  (armone.east) -- (armjoin |- armone.east);

\draw[semithick,draw=leanInk,densely dotted]
  (armtwo.east) -- (armjoin |- armtwo.east);

\draw[semithick,draw=leanInk,densely dotted]
  (armlast.east) -- (armjoin |- armlast.east) -- (armjoin);

\draw[flow,densely dotted]
  (armjoin) -- (fixed.west);

\draw[flow] (fixed.east) -- (initialize.west);

% Select a node, then generate its candidate tactics.
\node[box,text width=2.35cm,minimum height=1.85cm]
  (search) at (6.15,4.60)
  {\textbf{Best-first search}\\[3pt]
   Select a node\\
   with state $s$\\
   for expansion.};

\node[box,draw=leanBlue,fill=leanBlue!7,
      text width=2.75cm,minimum height=1.85cm]
  (model) at (9.60,4.60)
  {\textbf{Tactic generation}\\[3pt]
   Generate several\\
   candidate tactics.\\[3pt]
   Input: $s_0,\hat z,s$};

\draw[flow] (initialize.south) -- (13.30,6.15)
  -- (6.15,6.15) -- (search.north);

\draw[flow] (search.east) --
  node[note,above=3pt] {$s$}
  (model.west);

% Indented pseudocode inside the Lean-interaction box.
\node[box,inner sep=0pt,
      minimum width=4.40cm,minimum height=4.10cm]
  (lean) at (13.80,3.55) {};

\node[font=\sffamily\footnotesize\bfseries,text=leanInk]
  at (13.80,5.28) {Lean interaction};

\begin{scope}[xshift=-3pt,
  every node/.style={
    font=\sffamily\footnotesize,text=leanInk,
    anchor=base west,inner sep=0pt
  }
]
  % Indentation levels: x = 11.82, 12.12, 12.42.
  \node at (11.82,4.80)
    {\textit{For each} candidate tactic $a$:};

  \node at (12.12,4.44)
    {$s'=\Transition(s,a)$};

  \node at (12.12,4.08)
    {\textit{if} $s'=\bot$:};

  \node at (12.42,3.72)
    {discard candidate.};

  \node at (12.12,3.36)
    {\textit{else if} $s'=\top$:};

  \node (found) at (12.42,3.00)
    {solution found.};

  \node at (12.12,2.64)
    {\textit{else}:};

\node at (12.42,2.30)
  {add transition $s\xrightarrow{a}s'$};

\node at (12.42,1.96)
  {to the search graph.};

\node at (12.42,1.64)
  {\scriptsize (reuse nodes for duplicate states)};
\end{scope}

\draw[flow] (model.east) -- (lean.west |- model.east);

% Return to node selection only after checking all candidates.
\coordinate (resume) at (6.15,2.80);

\draw[flow] (lean.west |- resume) --
  node[note,above=1pt,inner sep=0pt]
    {No solution found\\Mark node with state $s$ as expanded}
  (resume) -- (search.south);

% Immediate exit from the successful case inside the Lean box.
\coordinate (solutionport) at ([xshift=3pt]found.east);
\fill[leanGreen] (solutionport) circle (1.2pt);

\node[box,draw=leanGreen,fill=leanGreen!10,
      text width=3.20cm,minimum height=.90cm]
  (success) at (13.80,.75)
  {\textbf{Return solution}\\
   Stop all arms.};

\draw[flow,draw=leanGreen]
  (solutionport) -- (16.08,0 |- solutionport)
  |- (success.east);

% Search exhaustion advances to the next arm.
\node[box,draw=leanOchre,fill=leanOchre!8,
      text width=4.65cm,minimum height=.90cm]
  (next) at (7.50,.75)
  {\textbf{Continue to the next arm}\\
   Return failure if no arms remain.};

\draw[flow] (search.west) -- (4.73,4.60) -- (4.73,1.65)
  -- node[note,above=1pt,inner sep=0pt,text width=2.65cm]
     {No nodes left to expand\\or budget exhausted}
  (7.50,1.65) -- (next.north);
\end{tikzpicture}%
}
\endgroup
\caption{Hierarchy-guided Lean inference with $J$ sequential arms.
Each arm samples a hierarchy from $F_\theta$ and subsequently invokes the solver with this fixed hierarchy and a fresh proof search graph rooted at the initial state $s_0$.
An unsuccessful arm is followed by the next and a valid solution stops the entire procedure; we signal failure if all $J$ arms have been exhausted.
The right panel illustrates tactic generation and Lean interaction within one arm.}
\label{fig:multi_arm_lean_search}
\end{figure}
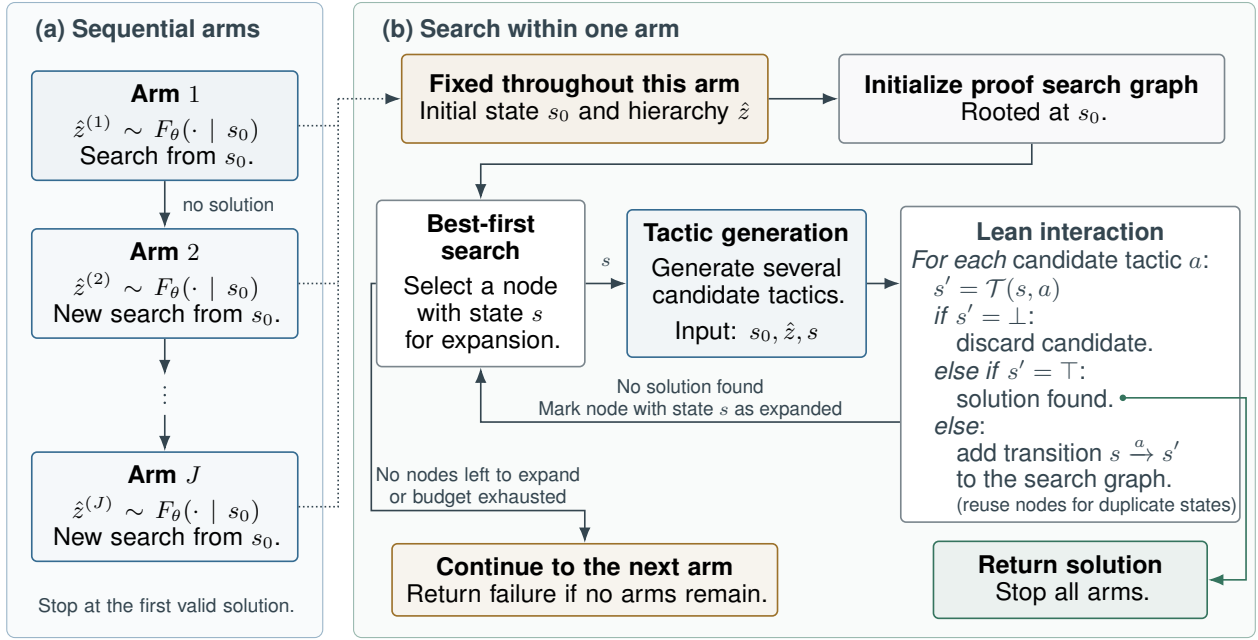

\subsection{Lean instantiation: $\ConstructTriple$}
\label{sec:lean_ConstructTriple}

We now specify concrete choices for the joint procedure $(\Utility,\DisSol)$ and the penalty $\Penalty$, which, together with Algorithm \ref{alg:verifier_guided_hindsight_hierarchy_selection}, determine $\ConstructTriple$ in our Lean instantiation.
These choices are not essential to the general method; we give them to provide a concrete instantiation.

\paragraph{Joint procedure $(\Utility , \DisSol )$.}
Usefulness can be assessed, for example, using the success ratio of repeated attempts under a prescribed proof search budget, optionally incorporating quantities such as consumed search resources and the lengths of discovered proofs. Our concrete choice below combines the success ratio with a preference for shorter proofs:

Given $(x,y,z)\in\X\times\Y\times\Z$ with $\Verify (x,y)=1$, we assess the usefulness of the hierarchy $z$ for finding a solution for $x$ by executing $N\in\mathbb{Z}_{\geq 1}$ runs of the solver component $S_\theta (\cdot\mid x,z)$ under a prescribed proof search budget.
Each run starts with a fresh proof search graph rooted at $x$, with the hierarchy $z$ and the parameters $\theta$ held fixed.
A run is successful if it returns a valid solution for $x$.

If none of these $N$ runs succeeds, we set
\begin{align*}
    (\Utility , \DisSol ) (x,y,z;\theta ) \leftarrow (0, y).
\end{align*}
Otherwise, denote the number of successful runs by
$M\in\{1,\dots,N\}$ and their returned solutions, in attempt order,
by $y_i\in\Actions^{k_i}$, $i=1,\dots,M$.
Note that, for $i\neq j$, the finite tactic sequences $y_i$ and $y_j$ need not be distinct.
We then choose a solution that minimizes proof length, i.e., number of tactics, among the discovered solutions and compute a usefulness estimate based on proof success ratio $r = M / N$ and average proof length $\ell = (k_1 + \dots + k_M) / M$.

More precisely, with $m\in\argmin_{i\in\{1,\dots,M\}}k_i$ (breaking ties by attempt order) and a hyperparameter $\eta>0$, we let
\begin{align*}
    (\Utility , \DisSol ) (x,y,z;\theta ) \leftarrow (r + \eta / \ell , y_m ).
\end{align*}

\paragraph{Penalty for hierarchies.}
In order to encourage the generation of hierarchies that represent proof ideas at different levels of granularity, we use a penalty that encourages compression between successive hierarchy levels. Concretely, with hyperparameters $\alpha_1 , \dots , \alpha_n \in (0,1)$ and the usual convention for empty sums (the depth $n$ may be $0$), we define
\begin{align*}
    \Penalty (z)
    =
    \sum_{j=1}^{n} \operatorname{ReLU}(\length (z_j)-\alpha_j\length (z_{j-1})) 
    ,
\end{align*}
for $z=(z_0 , \dots , z_n )\in \Z_0\times \cdots \times \Z_n$, where $\length (z_j )$ denotes the number of characters in the string $z_j$.
This choice of $\Penalty$ encourages compression, but does not by itself establish that the hierarchy is semantically meaningful.

\begin{remark}
With these choices, a hierarchy for which no solver run succeeds has $\Utility$ zero and hence a non-positive $\Score$ in the context of Algorithm \ref{alg:verifier_guided_hindsight_hierarchy_selection}.
While the fallback $y$ for $\DisSol$ makes the joint procedure well-defined in this case, Algorithm \ref{alg:verifier_guided_hindsight_hierarchy_selection} will not return the corresponding triple, since the hierarchy’s $\Score$ lies below the positive acceptance threshold $\delta$.
Consequently, every triple returned by Algorithm \ref{alg:verifier_guided_hindsight_hierarchy_selection} contains a valid solution \emph{discovered by the solver} under the triple's hierarchy.
\end{remark}

\subsection{Lean instantiation: training operator $\Train$}
\label{sec:lean_train}

We implement $\Train$ by supervised fine-tuning of the shared language model underlying the three system components.
We first describe how training triples are converted into prompt-response pairs, and then specify the composition of mini-batches and the training objective.

\paragraph{Prompt-response pairs.}
Given a dataset $D$ of triples $(x,y,z)\in\X\times\Y\times\Z$ with $\Verify (x,y)=1$, we proceed as follows.
If $D$ is empty, the parameters $\theta$ remain unchanged. Otherwise, let $(x,y,z)$ be an entry of $D$ and write $y=(a_1 , \dots , a_k )$, $s_0 = x$, and $s_{t+1}=\Transition (s_t , a_{t+1})$ for $t=0,\dots , k-1$.
In accordance with the description of the system components in Section \ref{sec:lean_system_components}, the triple $(x,y,z)$ yields a total of $2n+k+2$ prompt-response pairs, namely
\begin{itemize}
    \item $n+1$ prompt-response pairs for the foresight component: for predicting $z_n$ from $x$ and, for $j=0, \dots , n-1$, predicting $z_j$ from $(x, z_{j+1})$,
    \item $n+1$ prompt-response pairs for the hindsight component: for predicting $z_0$ from $(x, y)$ and, for $j=1, \dots , n$, predicting $z_j$ from $(x, z_{j-1})$,
    \item $k$ prompt-response pairs for the solver component: for predicting $a_{t+1}$ from $(x , z, s_t)$ for $t=0,\dots , k-1$,
\end{itemize}
where all occurring prompts use the prompt templates from Appendix \ref{sec:appendix_prompt_templates}. Note that, for the solver component, these pairs are intended to train the language model for tactic generation within best-first proof search.

\paragraph{Shuffling and balanced batching.}
For nonempty $D$, we group the constructed prompt-response pairs by component and, for foresight and hindsight, by hierarchy level.
Thus, we have one solver group and, for each $j=0 , \dots , n$,
one foresight group and one hindsight group, for a total of $2n+3$ groups.
Since a proof with $k$ tactics contributes $k$ solver pairs,
longer proofs contribute more entries within the solver group.
We fix positive integers $b_S , b_F , b_H$.
Each mini-batch contains $b_S$ pairs from the solver group, $b_F$ pairs from each foresight group, and $b_H$ pairs from each hindsight group.
Thus, each mini-batch contains a total of
\begin{align*}
    b = b_S + (n+1)(b_F + b_H )
\end{align*}
prompt-response pairs.
Here, a mini-batch denotes the full collection of prompt-response pairs used for one optimizer update.
Consequently,
\begin{align*}
    \beta_S
    =\frac{b_S}{b}
    ,
    &&
    \beta_F
    =\frac{(n+1)b_F}{b}
    ,
    &&
    \beta_H
    =\frac{(n+1)b_H}{b}
\end{align*}
are the fractions of pairs in each mini-batch belonging to the solver, foresight, and hindsight components, respectively.

Within each of the $2n+3$ groups, we shuffle the entries and traverse the resulting sequence in order; this shuffling happens independently between groups.
To construct a mini-batch, we take the next prescribed number
of entries from each group, see $b_S , b_F , b_H$ above, continuing from where the preceding mini-batch left off.
Whenever a group is exhausted, we reshuffle that group and
continue from the beginning of its new order, without resetting
the other groups.
This also applies if a group is exhausted
while a mini-batch is being constructed.

\paragraph{Training objective.}
Write $p_\theta$ for the conditional next-token probabilities
of the shared language model.
For a tokenized prompt $u=(u_1 , \dots ,u_m )$ and a tokenized
target response $v=(v_1 , \dots , v_L )$, \emph{in}cluding its
end-of-response marker but \emph{ex}cluding padding, we use the loss
\begin{align*}
    \ell_\theta (u,v)
    =
    -\frac{1}{L}
    \sum_{q=1}^{L}
    \log p_\theta (v_q \mid u, v_1 , \dots , v_{q-1})
    .
\end{align*}
Note that $L\geq 1$, since $v$ includes its end-of-response marker.
The normalization by $L$ was included, since hierarchy descriptions can be substantially longer than individual Lean tactics.

For nonempty $D$, let $\Loss_S (\theta ;D)$ denote the arithmetic mean of $\ell_\theta$ over all solver prompt-response pairs constructed from $D$.
Similarly, for $j=0,\dots , n$, let $\Loss_{F,j} (\theta;D)$ and $\Loss_{H,j} (\theta;D)$ denote the arithmetic mean of $\ell_\theta$ over all foresight and hindsight prompt-response pairs at hierarchy level $j$, respectively.
These averages count repeated entries separately.
We train the shared language model underlying the three components using mini-batch optimization of the objective
\begin{align*}
    \Loss (\theta ;D)
    =
    \beta_S\Loss_S (\theta ;D)
    +\frac{\beta_F}{n+1}
      \sum_{j=0}^{n}\Loss_{F,j}(\theta ;D)
    +\frac{\beta_H}{n+1}
      \sum_{j=0}^{n}\Loss_{H,j}(\theta ;D),
\end{align*}
where $\beta_S , \beta_F , \beta_H$ are the fractions for the three components per mini-batch introduced in the preceding paragraph.

The procedure $\Train$ is then implemented by iterating over mini-batches and executing one optimizer step per mini-batch, where the loss used for an optimizer step is the arithmetic mean of $\ell_\theta$ over the prompt-response pairs in the mini-batch.
Concrete batch counts, optimizer settings, and training duration are specified in Appendix \ref{sec:appendix_Train}.

\section{Discussion}
\label{sec:discussion}

\paragraph{From hindsight to foresight.}
Our central hypothesis is that reverse engineering the idea underlying a known solution is, in general, easier than discovering it from the problem alone.
We operationalize this hypothesis through a training loop that alternates between reverse engineering ideas from known solutions (hindsight) and training the model using these ideas.
When reverse engineering solution ideas, it is crucial to avoid hindsight bias.
In our implementation, we seek to mitigate this risk through corresponding instructions in the language model's prompt templates.
Nevertheless, it is a priori not clear that solution ideas derived in hindsight are suitable targets for foresight prediction.
This highlights the need for a thorough experimental evaluation of the proposed method.

\paragraph{The case of verifiable problem domains.}
In verifiable problem domains we have access to a verifier that either certifies or rejects proposed solutions.
We exploit this by refining the hindsight idea generation in the alternating training loop: Given a problem and a reference solution, we generate several ideas in hindsight and ask the model to solve the problem using the generated ideas.
We then use the verifier and domain-specific quantities to evaluate the proposed solutions and subsequently score the usefulness of the generated ideas.
The highest-scoring idea is retained (provided a certain threshold is met), while the other ideas are discarded.
This introduces an additional training signal in the spirit of expert iteration.
However, a high score does not by itself establish that the retained idea helped the model solve the problem.
For example, when not orchestrating the training loop carefully, the model might learn to ignore the proposed ideas and generate solutions based on the problem alone.
This once again highlights the need for a thorough evaluation of the proposed method, including ablation studies assessing whether selecting hindsight-generated ideas based on score improves model performance.

\paragraph{Future empirical validation.}
Future work should experimentally evaluate the proposed method.
The evaluation should cover both the Lean instantiation and the version without a verifier, and include comparisons with standard supervised fine-tuning under comparable data and computational budgets, as well as thorough ablation studies.

\section*{Use of AI tools}

The method and its original rigorous formal specification (i.e., the content of Sections \ref{sec:problem}, \ref{sec:method}, and \ref{sec:concrete_lean_instantiation}) were developed without AI assistance.
ChatGPT, using multiple model versions, assisted with the preparation of this manuscript.
This assistance included drafting and revising prose from author-provided technical material, as well as  searching for and discussing related literature.
It also included assistance with generating figures, \LaTeX{} preparation, and typesetting.

AI assistance was also used for implementation and testing in the associated software project.
This version of the manuscript reports no empirical evaluation.
The human authors retain full responsibility for the manuscript, including its scientific claims, mathematical content, references, and all material prepared with AI assistance.

\bibliographystyle{ieeetr}
\bibliography{references}

\appendix
\section{Appendix}
\label{sec:appendix}

This appendix provides the details for our implementation of the Lean instantiation of the proposed method.

\subsection{Language model and environment}
\label{sec:appendix_model}

\begin{center}
\small
\begin{tabular}{@{}ll@{}}
Shared model & \texttt{internlm/internlm2-math-plus-7b} \citep{internlm2024plus} \\
Weights/tokenizer revision & \texttt{2565e424cc6f83358e7a5aa19b639205fcdd173d} \\
Weights; context & bfloat16; 8192 tokens \\
Hierarchy depth & $n=1$ \\
Theorem-proving environment & Lean 4.10.0 \citep{lean42021}; LeanDojo 2.1.3 \citep{leandojo2023} \\
Runtime & Python 3.11.9; x86\_64 Linux \\
Reference inference packages & vLLM 0.4.1; PyTorch 2.2.1; Transformers 4.40.1
\end{tabular}
\end{center}
All trainable parameters are shared; prompts distinguish components.

\subsection{Data}
\label{sec:appendix_data}

\paragraph{External data and seed split.}
We extract named theorems from mathlib \citep{mathlib2020}, revision
\begin{center}
\small\texttt{a719ba5c3115d47b68bf0497a9dd1bcbb21ea663}.
\end{center}
Following the random splitting procedure of LeanDojo \citep{leandojo2023}, we partition the extracted theorems into training, validation, and test sets, reserving 2,000 theorems each for validation and testing (random seed 3407).
Only the training partition is used to construct the seed
dataset and obtain subsequent training batches.
From the training partition, we retain only complete tactic proofs that, when executed in Lean, close all goals and reproduce the proof states recorded in the dataset.
Proofs with missing tactics are excluded.
This yields 50,523 pairs $(x,y)$ with $\Verify(x,y)=1$.
We randomly permute these pairs using seed 20260523, assign the first 15,000 to seed hierarchy generation, and reserve the remaining 35,523 for subsequent batches.
Both subsets are used for training; neither is an evaluation set.

\paragraph{Constructing the seed dataset: teacher hierarchies.}
The teacher language model, \texttt{gpt-5.4-mini} (reasoning effort \texttt{high}, verbosity \texttt{low}), receives the initial state, complete tactic sequence, and theorem metadata, excluding intermediate states.
Its prompt (Appendix \ref{sec:appendix_prompt_templates}) requests a detailed plan $z_0$, a coarse idea $z_1$, and a short anti-hindsight-bias check.
We form $D_0$ by augmenting the 15,000 pairs assigned to seed
hierarchy generation with their corresponding hierarchies $(z_0,z_1)$.
Later hierarchies intended for training come exclusively from the current hindsight component.

\paragraph{Choice of $\Batch$.}
We propose processing the reserved pool in the order induced by the random permutation above, taking successive batches of up to 2,048 entries.
After processing the entire pool, we revisit pairs for which $\ConstructTriple$ returned $\fail$; we allow at most two such retry passes, as specified in Appendix \ref{sec:appendix_hierarchy_selection}.
This uses neither synthetic data nor difficulty groups (Section \ref{sec:ingredients}).
The associated replay policy and training duration are specified in Appendices \ref{sec:appendix_replay} and \ref{sec:appendix_Train}.

\subsection{Role-specific prompt templates}
\label{sec:appendix_prompt_templates}

For $n=1$ (Section \ref{sec:lean_hierarchies_and_data}), $z_0$ is labelled \texttt{DETAILED PLAN} and $z_1$ \texttt{COARSE IDEA}; angle brackets mark inserted text. The teacher language model produces both levels per response; each shared-model foresight or hindsight call produces one.

\paragraph{Teacher language model.}
The teacher receives the initial state, verified tactics, and theorem metadata, without intermediate states. Its JSON response must contain exactly the three string fields below. Only the first two are hierarchy targets; \texttt{anti\_hindsight\_check} records an explanation.

\begin{quote}
\small\ttfamily
You are helping create training data for a Lean theorem-proving system.

For each verified Lean theorem proof, write two forward-usable proof ideas that describe the same proof strategy at different levels of detail:

1. a detailed proof plan, and\newline
2. a shorter, more abstract proof idea.

You will see:

1. the original Lean 4 proof state before any tactic is run, and\newline
2. a verified tactic proof that solves the theorem.

Your task is to infer the mathematical proof idea behind the verified proof and rewrite it as guidance that could plausibly have been used from the original state.

Critical anti-hindsight-bias rule:

- The verified proof is evidence for inferring the idea, not something to copy.\newline
- Do not write a chronological replay of the tactic proof.\newline
- Do not say "the proof does X", "the verified proof uses X", "after seeing the proof", or similar hindsight-only language.\newline
- Do not rely on information that is only visible after running later tactics, unless you phrase it as a forward-usable mathematical reason already suggested by the initial state.\newline
- Tactic names may be mentioned only when they express a reusable strategy visible from the initial state, such as rewriting a definition, splitting cases, applying induction, using a known hypothesis, or discharging arithmetic.

Return only valid JSON with this exact shape:

\textasciigrave{}\textasciigrave{}\textasciigrave{}json\newline
\{\newline
"z0\_detailed\_plan": "A forward-usable detailed proof plan. It may mention the useful definitions, hypotheses, transformations, and proof obligations, but it must not be a tactic-by-tactic replay.",\newline
"z1\_coarse\_idea": "A short reusable proof idea at a higher level of abstraction.",\newline
"anti\_hindsight\_check": "One sentence explaining why the plan could plausibly be chosen from the initial state alone."\newline
\}\newline
\textasciigrave{}\textasciigrave{}\textasciigrave{}

Requirements for \textasciigrave{}z0\_detailed\_plan\textasciigrave{}:

- 2 to 6 sentences.\newline
- Explain why the strategy should work mathematically.\newline
- Mention relevant hypotheses, definitions, lemmas, or theorems when they are visible from the initial state or are natural standard tools for the theorem's mathematical context.\newline
- Avoid copying the tactic sequence.

Requirements for \textasciigrave{}z1\_coarse\_idea\textasciigrave{}:

- 1 sentence.\newline
- More abstract and shorter than \textasciigrave{}z0\_detailed\_plan\textasciigrave{}.\newline
- Reusable across similar theorem states.

Initial Lean 4 state:

\textasciigrave{}\textasciigrave{}\textasciigrave{}lean\newline
\textless{}INITIAL\_STATE\textgreater{}\newline
\textasciigrave{}\textasciigrave{}\textasciigrave{}

Verified tactic proof:

\textasciigrave{}\textasciigrave{}\textasciigrave{}lean\newline
\textless{}TACTIC\_PROOF\textgreater{}\newline
\textasciigrave{}\textasciigrave{}\textasciigrave{}

Theorem metadata:

\textasciigrave{}\textasciigrave{}\textasciigrave{}text\newline
theorem\_full\_name: \textless{}THEOREM\_FULL\_NAME\textgreater{}\newline
file\_path: \textless{}FILE\_PATH\textgreater{}\newline
proof\_step\_count: \textless{}PROOF\_STEP\_COUNT\textgreater{}\newline
\textasciigrave{}\textasciigrave{}\textasciigrave{}
\end{quote}

The anti-hindsight-bias instruction encourages guidance usable from the initial state.

\paragraph{Shared-model message format.}
Foresight, hindsight, and tactic generation use the same underlying language model, initialized from \texttt{internlm/internlm2-math-plus-7b} (Appendix \ref{sec:appendix_model}) and jointly fine-tuned as described in Section \ref{sec:lean_train}.
Their prompts use the following chat wrapper, matching the chosen model's chat format, without a system message:

\begin{quote}
\par\nopagebreak\noindent\begin{minipage}{\linewidth}
\small
\begin{verbatim}
<|im_start|>user
<USER MESSAGE><|im_end|>
<|im_start|>assistant
\end{verbatim}
\end{minipage}\par
\end{quote}

Each template below specifies a complete user message for the
chat wrapper above.
Responses end with \texttt{<|im\_end|>}.
Displayed line breaks within prose sentences serve typesetting only.

\paragraph{Foresight: coarse idea, $F_{1,\theta}$.}
Request $z_1$ from the initial state:

\begin{quote}
\par\nopagebreak\noindent\begin{minipage}{\linewidth}
\small
\begin{verbatim}
My initial LEAN 4 state is:
```lean
<INITIAL STATE>
```

Please write a concise proof idea that could help prove the theorem
from this state.
Write only the requested proof idea, with no extra commentary.
COARSE IDEA:
\end{verbatim}
\end{minipage}\par
\end{quote}

\paragraph{Foresight: detailed plan, $F_{0,\theta}$.}
Request $z_0$ from the initial state and the preceding coarse idea:

\begin{quote}
\par\nopagebreak\noindent\begin{minipage}{\linewidth}
\small
\begin{verbatim}
My initial LEAN 4 state is:
```lean
<INITIAL STATE>
```

Here is the previous proof idea:
COARSE IDEA:
<COARSE IDEA>

Please refine the previous proof idea into a more detailed proof idea
that could help prove the theorem from this state.
Write only the requested proof idea, with no extra commentary.
DETAILED PLAN:
\end{verbatim}
\end{minipage}\par
\end{quote}

\paragraph{Hindsight: detailed plan, $H_{0,\theta}$.}
Request $z_0$ from the initial state and the reference tactic sequence,
with tactics separated by newlines:

\begin{quote}
\par\nopagebreak\noindent\begin{minipage}{\linewidth}
\small
\begin{verbatim}
My initial LEAN 4 state is:
```lean
<INITIAL STATE>
```

Here is the verified Lean 4 tactic proof:
```lean
<TACTIC SEQUENCE>
```

Reverse-engineer a detailed proof idea from this verified proof.
Avoid hindsight bias: write a forward-usable idea that could plausibly
have been followed from the original state, not an idea that can only
be obtained from knowledge of the complete proof.
Write only the requested proof idea, with no extra commentary.
REVERSE ENGINEERED DETAILED PROOF IDEA:
\end{verbatim}
\end{minipage}\par
\end{quote}

\paragraph{Hindsight: coarse idea, $H_{1,\theta}$.}
Request $z_1$ from the initial state and the preceding detailed plan:

\begin{quote}
\par\nopagebreak\noindent\begin{minipage}{\linewidth}
\small
\begin{verbatim}
My initial LEAN 4 state is:
```lean
<INITIAL STATE>
```

Here is the previous, more detailed proof idea:
DETAILED PLAN:
<DETAILED PLAN>

Summarize the previous proof idea into a coarser forward-usable proof
idea. Preserve the strategic reason it could help from the initial
state.
Write only the requested proof idea, with no extra commentary.
SUMMARIZED IDEA:
\end{verbatim}
\end{minipage}\par
\end{quote}

\paragraph{Solver: next tactic.}
Request a tactic from the initial state, the hierarchy, and the
current proof state.
The prompt explicitly allows the hierarchy to be imperfect:

\begin{quote}
\par\nopagebreak\noindent\begin{minipage}{\linewidth}
\small
\begin{verbatim}
My initial LEAN 4 state is:
```lean
<INITIAL STATE>
```

Here is a suggested proof idea at several levels of detail. It may be
useful, incomplete, or partly wrong. Use it as guidance where it fits
the current LEAN 4 state, but choose the next tactic based on the
current state.

COARSE IDEA:
<COARSE IDEA>

DETAILED PLAN:
<DETAILED PLAN>

My current LEAN 4 state is:
```lean
<CURRENT STATE>
```
Please predict a possible tactic to help me prove the theorem.
\end{verbatim}
\end{minipage}\par
\end{quote}

Only the current state changes during search; the initial state and
hierarchy remain fixed.
The foresight, hindsight, and solver templates above also define the
prompts of the supervised prompt-response pairs described in
Section \ref{sec:lean_train}.

\paragraph{Null arm.}
The optional null arm (Appendix \ref{sec:appendix_multiple_arms})
makes no foresight call.
It uses the solver template with both hierarchy placeholders replaced
by \texttt{No proof idea provided.}, yielding the following user message:

\begin{quote}
\par\nopagebreak\noindent\begin{minipage}{\linewidth}
\small
\begin{verbatim}
My initial LEAN 4 state is:
```lean
<INITIAL STATE>
```

Here is a suggested proof idea at several levels of detail. It may be
useful, incomplete, or partly wrong. Use it as guidance where it fits
the current LEAN 4 state, but choose the next tactic based on the
current state.

COARSE IDEA:
No proof idea provided.

DETAILED PLAN:
No proof idea provided.

My current LEAN 4 state is:
```lean
<CURRENT STATE>
```
Please predict a possible tactic to help me prove the theorem.
\end{verbatim}
\end{minipage}\par
\end{quote}

\subsection{Best-first proof search configuration}
\label{sec:appendix_best_first_search}

The following settings apply to inference; hierarchy assessment during training uses smaller limits, see Appendix \ref{sec:appendix_hierarchy_selection}.
\begin{center}
\small
\begin{tabular}{@{}p{0.31\linewidth}p{0.64\linewidth}@{}}
Tactic decoding & Deterministic beam search; 32 responses; $\leq256$ new tokens each; length penalty $0$; stop \texttt{<|im\_end|>}. \\
Per-attempt search limits & 100 node expansions; depth 100; 32 tactics per expanded state; nominal time 600 seconds. \\
LeanDojo tactic timeout & 300 seconds. \\
\end{tabular}
\end{center}
The parser removes tactic wrappers and rejects empty outputs; duplicate tactic texts are removed before execution.
While beam search generates candidate tactics for the selected proof state, the priority rule below determines which node in the proof search graph to expand next.

\paragraph{Search priority.}
A newly created node in the proof search graph is assigned the priority
\begin{align*}
    P(a_1 ,\dots ,a_d ) = \sum_{t=1}^{d}\frac{g_t}{c_t},
\end{align*}
where $(a_1 ,\dots ,a_d )$ is the sequence of tactics along the path by which the node was first reached from the root.
For each $t=1,\dots ,d$, $c_t\geq 1$ is the number of tokens in the response generated by the language model from which tactic $a_t$ was extracted, and $g_t$ is the sum of their conditional log probabilities under the language model.
Each token probability is conditioned on the solver prompt used at that step and all preceding tokens of that response.
These quantities refer to the generated response before
removing tactic wrappers; prompt tokens are not counted.
The root node has priority zero.

\paragraph{State reuse and termination.}
At each expansion, select the unexpanded node in the queue with the highest priority, breaking ties by insertion order.
For the selected node, execute the candidate tactics generated by the language model in Lean in decreasing order of their step scores, where a step score is the average conditional log probability over the tokens of the corresponding generated response.
Each tactic is applied independently to the same proof
state associated with the selected node.
Discard failed tactics; if a tactic completes the proof, return the discovered solution without executing the remaining candidate tactics.
If a tactic produces a valid successor state with open goals, record the corresponding state and transition in the proof search graph: If the state has occurred before, use the corresponding existing node (without updating its depth and priority), otherwise create a new node, compute its search priority, and insert it into the queue; in either case, add an edge to this node from the selected node, labelled by the tactic.
If none of the candidate tactics completes a proof, mark the selected node as expanded.
Expanded nodes which receive a new incoming edge retain their expanded status.
Nodes at the depth limit are marked as expanded without generating candidate tactics.
Treat a tactic timeout as a failed tactic and continue with
the remaining candidates.
Return failure when no nodes remain to expand or the overall search budget is exhausted.

\subsection{Inference: multiple arms}
\label{sec:appendix_multiple_arms}

For the sequential procedure in Section \ref{sec:lean_system_components} (see also Figure \ref{fig:multi_arm_lean_search}), we choose $J=2$ hierarchy-conditioned arms. Each generates a new hierarchy and subsequently invokes the solver component with the full budget from Appendix \ref{sec:appendix_best_first_search}. The parameters $\theta$ of the shared language model stay fixed and we stop at the first valid solution.

Foresight and hindsight use the same per-level decoding:
\begin{center}
\small
\begin{tabular}{@{}p{0.31\linewidth}p{0.64\linewidth}@{}}
Sampling & One response; temperature $0.7$; nucleus threshold $0.95$. \\
Length and termination & $\leq512$ new tokens; stop \texttt{<|im\_end|>}. \\
Foresight seed schedule & First arm: 20260820, 20260821; second: 20260822, 20260823, in generation order. \\
\end{tabular}
\end{center}

\paragraph{Optional null arm.}
We may optionally prepend a \emph{null arm} to the hierarchy-conditioned arms described in Section \ref{sec:lean_system_components}: Such an arm skips foresight hierarchy generation and calls the solver component with the problem $x$ and the hierarchy $(z_0 , z_1 )=(\texttt{No proof idea provided.},\texttt{No proof idea provided.})$; the corresponding prompt template for the shared language model can be found in Appendix \ref{sec:appendix_prompt_templates}. It uses the same budget and proof search configuration as regular hierarchy-conditioned arms, see Appendix \ref{sec:appendix_best_first_search}.

\subsection{Hierarchy selection and iteration settings}
\label{sec:appendix_hierarchy_selection}

We propose the following configuration for implementing the Lean instantiation of $\ConstructTriple$, see Section \ref{sec:lean_ConstructTriple}, and the full training loop. These settings specify a concrete starting point for evaluation; they have not been selected through empirical comparison.
\begin{center}
\small
\begin{tabular}{@{}p{0.31\linewidth}p{0.64\linewidth}@{}}
Candidates per pair & $K=4$ hindsight hierarchies, each from separate $H_{0,\theta}$ then $H_{1,\theta}$ calls. \\
Candidate decoding & Temperature $0.7$; nucleus threshold $0.95$; $\leq512$ new tokens per level; fresh pseudorandom seed per call, recorded with the hierarchy. \\
Assessment & $N=1$ solver run per hierarchy; 20 node expansions; nominal time 120 seconds; LeanDojo tactic timeout 30 seconds. Other settings as in Appendix \ref{sec:appendix_best_first_search}. \\
Selection coefficients & $\eta=1$, $\alpha_1=\tfrac12$, $\lambda=10^{-3}$, $\delta=\tfrac12$. \\
Pass limit & One complete pass through the 35,523 reserved pairs, followed by at most two retry passes. \\
\end{tabular}
\end{center}
Deterministic tactic beam search makes repeated runs with the same hierarchy redundant; the general definition (Section \ref{sec:lean_ConstructTriple}) allows $N>1$, which makes sense when using non-deterministic decoding strategies for tactic generation or other search randomization.
To limit the cost of hierarchy assessment during the full training loop, we use smaller proof search budgets than during inference (see Appendix \ref{sec:appendix_best_first_search}).
All candidates use the same assessment limits.

Success with $k$ tactics gives usefulness $1+1/k$; failure gives zero. We measure the penalty in characters of the decoded hierarchy strings.

\paragraph{Passes and retries.}
We iterate over queues of pairs, processing their entries in batches of up to 2,048, and initialize the first queue with the reserved pairs in the order from Appendix \ref{sec:appendix_data}.
Within each pass through a queue, we append pairs for which $\ConstructTriple$ returns $\fail$ to a separate retry queue, preserving their order.
After the current queue has been processed in full, the constructed retry queue becomes the queue for the next pass.
Pairs yielding an accepted triple leave the queue permanently; their triples remain available for replay.
A pair also enters the retry queue when a solution was found during assessment but no candidate hierarchy attained the acceptance threshold.

We stop if the retry queue is empty, a complete pass through a queue produced no accepted triple, or three passes have been completed.

The initial pass comprises 17 batches of 2,048 and one of 707 entries; each retry pass has at most 18 batches.
Thus, in Algorithm \ref{alg:training-loop}, take $T=54$ and let $\Batch(t)$ return an empty batch after one of the above stopping conditions is met; the resulting remaining iterations make no parameter updates and may be omitted in implementation.

\subsection{Replay policy}
\label{sec:appendix_replay}

For the proposed schedule, let $m_t$ denote the number of accepted triples in $D_t$.
If $m_t>0$, sample $m_t$ entries uniformly without replacement from the concatenation of $D_0,\dots,D_{t-1}$ and concatenate them with all entries of $D_t$ to form $D$.
Thus, $D$ contains $2m_t$ triples: equal numbers of new and replayed entries.
The history contains at least the 15,000 seed triples from $D_0$, while $m_t\leq2{,}048$, so this sampling is always possible in the concrete configuration.
If $m_t=0$, choose $D$ empty; $\Train$ then leaves the parameters unchanged.

\subsection{Implementation details for $\Train$}
\label{sec:appendix_Train}

We use the five groups and objective of Section \ref{sec:lean_train} with $n=1$ and the prompt templates in Appendix \ref{sec:appendix_prompt_templates}.
The following are proposed settings that remain to be evaluated empirically.

Let $|D|$ count dataset entries and let $N_S(D)$ denote the number of solver prompt-response pairs constructed from $D$.
For nonempty $D$, perform
\begin{align*}
    U(D)=\max\left\{
        \left\lceil\frac{|D|}{2}\right\rceil,
        \left\lceil\frac{N_S(D)}{8}\right\rceil
    \right\}
\end{align*}
optimizer updates, traversing every group at least once (since $b_S = 8, b_F = b_ H = 2$, see below).
For our seed dataset, $N_S(D_0)=42{,}397$ and $U(D_0)=7500$.

\begin{center}
\small
\begin{tabular}{@{}p{0.31\linewidth}p{0.64\linewidth}@{}}
Mini-batch counts
& $b_S=8$, $b_F=b_H=2$ per level; $b=16$. \\
Optimizer
& AdamW; all shared parameters; otherwise default settings. \\
Peak learning rate
& $5\cdot10^{-6}$ for seed training; $10^{-6}$ subsequently. \\
Schedule
& $\lfloor U(D)/15\rfloor$ linear warmup steps,
then linear decay to zero. \\
Weight decay; gradient norm clip
& $0$; $1$. \\
Arithmetic; memory
& bfloat16; gradient checkpointing. \\
Shuffle/training seeds
& 20260609 for seed training; $20260609+t$ in iteration $t$. \\
Maximum prompt+response length
& 8192 tokens.
\end{tabular}
\end{center}

Each call starts from the current parameters with
a fresh optimizer and freshly shuffled groups.
Empty $D$ leaves the parameters unchanged.
Tokenize prompts and responses separately without adding implicit special tokens, then concatenate them; responses include \texttt{<|im\_end|>}.
Every pair must fit within the context limit; no truncation
is applied.

\end{document}